\documentclass[11pt]{article}

\usepackage[preprint]{acl}

\usepackage{float}
\usepackage{enumitem}
\usepackage{subcaption}   
\usepackage{times}
\usepackage{multirow}
\usepackage{latexsym}
\usepackage{amsmath}
\usepackage{booktabs}
\usepackage{graphicx}
\usepackage{url}
\usepackage{stfloats}
\usepackage{booktabs}
\usepackage{tabularx}

\usepackage[T1]{fontenc}

\usepackage[utf8]{inputenc}

\usepackage{microtype}

\usepackage{inconsolata}

\usepackage[prependcaption,textsize=tiny]{todonotes}

\title{Can We Trust LLM Detectors?}

\author{
Jivnesh Sandhan\textsuperscript{1}, 
  Harshit Jaiswal\textsuperscript{2}, 
  Fei Cheng\textsuperscript{1} and
  Yugo Murawaki\textsuperscript{1}
  \\
 \textsuperscript{1}Kyoto University, Japan and \textsuperscript{2}IIT Kanpur, India. \\
 \texttt{\{jivnesh,feicheng,murawaki\}@i.kyoto-u.ac.jp} and \texttt{harshitj23@iitk.ac.in}
}

\begin{document}
\maketitle
\begin{abstract}
The rapid adoption of LLMs has increased the need for reliable AI text detection, yet existing detectors often fail outside controlled benchmarks. We systematically evaluate 2 dominant paradigms (training-free and supervised) and show that both are brittle under distribution shift, unseen generators, and simple stylistic perturbations. To address these limitations, we propose a supervised contrastive learning (SCL) framework that learns discriminative style embeddings. Experiments show that while supervised detectors excel in-domain, they degrade sharply out-of-domain, and training-free methods remain highly sensitive to proxy choice. Overall, our results expose fundamental challenges in building domain-agnostic detectors.
\footnote{Our code is available at: \url{https://github.com/HARSHITJAIS14/DetectAI}}
\end{abstract}

\section{Introduction}

The rapid adoption of large language models (LLMs) has intensified the need for reliable LLM detection, particularly in academic settings where pedagogical integrity and research authenticity are critical \cite{Orenstrakh2024}. While many existing detectors report near-perfect in-domain accuracy, their reliability in real-world, out-of-distribution (OOD) settings remains questionable \cite{sadasivan2023can}. This mismatch between benchmark performance and practical robustness undermines trust in automated detection systems. In this work, we systematically evaluate the reliability of 2 dominant detection paradigms (training-free and supervised fine-tuning) and propose a new framework to mitigate their OOD failures.

Current LLM detectors mainly follow 2 approaches. Supervised classifiers, often based on fine-tuned models such as BERT \cite{devlin2019bert}, achieve high in-domain accuracy but tend to overfit to model-specific artifacts, leading to severe degradation on text from unseen generators. In contrast, training-free detectors, including DetectGPT \cite{mitchell2023detectgpt}, Fast-DetectGPT \cite{bao2024fastdetectgpt}, and Binoculars \cite{hans2024spotting}, rely on intrinsic statistical properties of generated text. Although appealing, their performance is highly sensitive to the choice of proxy model, introducing a distinct but equally problematic form of brittleness.
\begin{figure*}[t]
    \centering
    \includegraphics[width=0.8\textwidth]{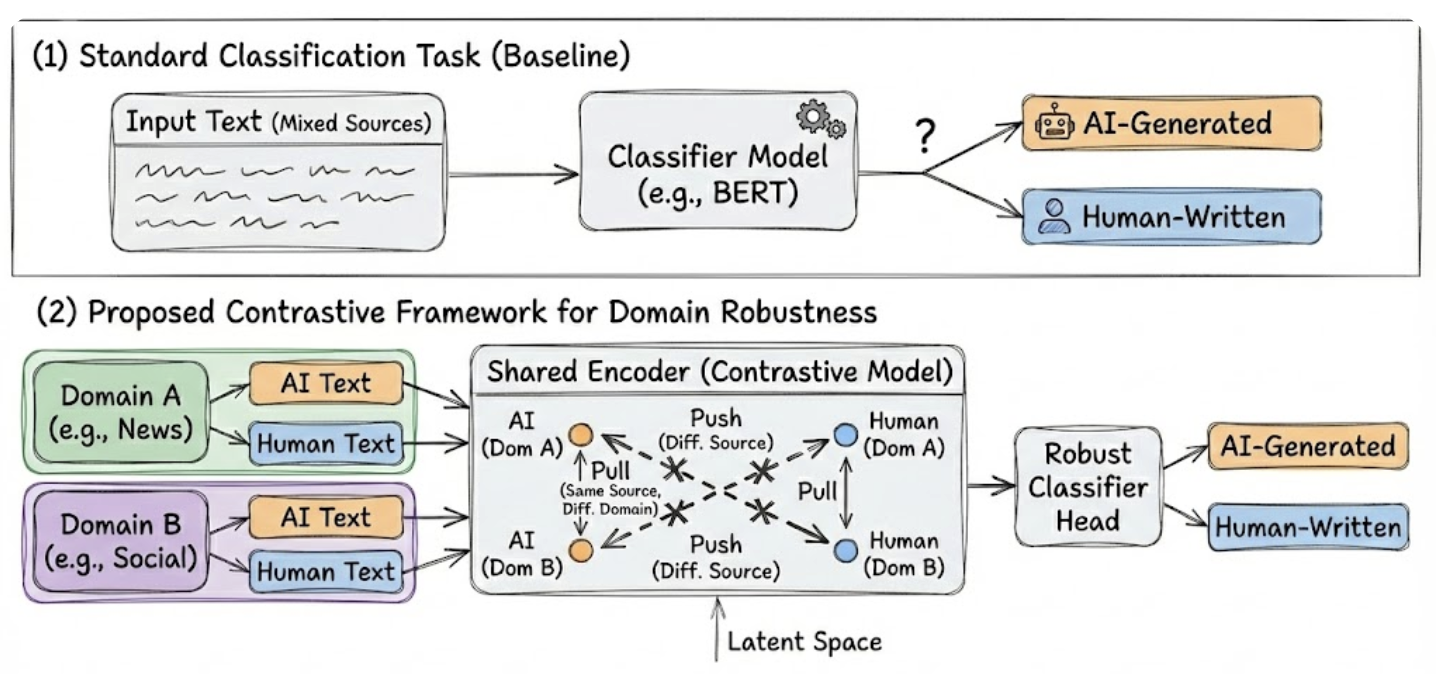}
    \caption{Overview of the AI text detection task and proposed framework. Our framework integrates a supervised classifier with a supervised contrastive learning module that learns style embeddings, enabling robust detection. }
    \label{fig:main}
\end{figure*}
To address these limitations, we propose a supervised contrastive learning (SCL) framework that learns discriminative style embeddings separating human and machine-generated text. Using a DeBERTa-v3 backbone optimized with an InfoNCE loss, our method enables efficient few-shot adaptation to new LLMs with as few as 25 examples. Extensive experiments show that supervised detectors degrade substantially under distribution shift, while training-free methods remain fragile to proxy selection. We find that limited supervision improves generalization but does not fully resolve these issues. Our robustness analysis further reveals vulnerabilities to adversarial attacks and simple stylistic perturbations, highlighting severity of reliability these detectors. Our contributions are:
\begin{itemize}[itemsep=0pt, parsep=0pt]
    \item We show that both training-free and supervised detectors fail under domain shift.
    \item We introduce a contrastive learning framework that improves robustness and enables few-shot adaptation to new LLMs.
    \item We provide comprehensive OOD and adversarial analyses, providing evidence that it is not possible to built universal detector.
\end{itemize}

\section{The proposed framework}
\label{proposed_method}
We use a pre-trained BERT model as our base model fine-tuned  with binary classification objective. The model is trained by minimizing the binary cross-entropy loss function:
\begin{equation}
\mathcal{L}_{\mathrm{BCE}} = -\left[ y \log(p) + (1 - y)\log(1 - p) \right]
\end{equation}

where $y$ is the true label and $p$ is the model's predicted probability. We propose a Supervised Contrastive Learning (SCL) module illustrated in Figure \ref{fig:main}, inspired by recent successes in learning discriminative representations \cite{guo2024detective, chen2024spcontrastnet, aberdam2021sequence}. This module helps learning a meaningful style embeddings using a DeBERTa-v3 backbone and a projection head. It is trained with the InfoNCE loss function:
\begin{equation}
\mathcal{L}_{\mathrm{InfoNCE}}
= -\log \frac{\exp\!\left(\operatorname{sim}(z_{i}, z_{j}^{+}) / \tau\right)}
{\sum_{k=1}^{N} \exp\!\left(\operatorname{sim}(z_{i}, z_{k}) / \tau\right)}
\end{equation}
where $z_i$ is an anchor embedding, $z_{j}^{+}$ is a positive sample, $\operatorname{sim}(\cdot,\cdot)$ is cosine similarity, and $\tau$ is a temperature parameter.  This objective structures the embedding space such that classes form distinct, coherent clusters.

\section{Experiments}
\paragraph{Datasets:} We use 3 datasets: \texttt{CHEAT}, with 35,304 academic abstracts for high-performance supervised evaluation \cite{zhao2023cheat}; \texttt{RAID}, a large-scale benchmark with over 6 million texts from 11 LLMs across 8 domains for cross-domain comparison \cite{dugan2024raid}; and \texttt{M4}, a multi-generator, multi-domain, and multi-lingual corpus designed for black-box detection that encompasses diverse linguistic settings and generative models to test detector generalization \cite{wang2024m4}.

\paragraph{Baselines:} We evaluate our method against two training-free and two supervised baselines. We use NVIDIA A100 and TITAN RTX GPUs with a learning rate of 2e-5 for models training.
\begin{itemize}[itemsep=0pt, parsep=0pt]
    \item \textbf{Binocular:} A perplexity-based, training-free detector compares token-level likelihoods under a base language model and its perturbed variant to find statistical regularities characteristic of AI-generated text \cite{hans2024spotting}.
    
    \item \textbf{FastDetectGPT:} A discrepancy-based, training-free approach that measures the divergence between the likelihood of a text under an original model and samples generated from it \cite{bao2024fastdetectgpt}.
    
    \item \textbf{BERT:} A supervised BERT-base classifier fine-tuned for binary classification.

   \item \textbf{GAN-BERT:}   We adopt GAN-BERT \cite{croce2020ganbert}, which integrates adversarial learning with supervised text classification. The model comprises a generator and a discriminator: the generator maps Gaussian noise $z \sim \mathcal{N}(0,1)$ to synthetic embeddings that mimic BERT representations, while the discriminator receives either real BERT embeddings or generated ones and predicts $k+1$ classes (the $k$ task labels plus a ``fake'' class). 
    \item \textbf{Ours:} The proposed method describe in \S \ref{proposed_method}.

\end{itemize}

\paragraph{Main Results:}
\begin{table*}[t]
\centering
\resizebox{0.6\textwidth}{!}{%
\begin{tabular}{@{}llccccc@{}}
\toprule
\textbf{Dataset} & \textbf{Method} & \textbf{Acc.} & \textbf{Prec.} & \textbf{Recall} & \textbf{F1} & \textbf{FPR} \\
\midrule
\multirow{5}{*}{\textbf{RAID}} 
&FastDetectGPT   & 34.63\% & 92.00\% & 2.00\%  & 4.00\%  & \textbf{0.35}\% \\
&Binoculars      & 82.81\% & 93.61\% & 79.65\% & 86.07\% & 10.87\% \\
&GAN-BERT        & 95.72\% & 95.85\% & 97.81\% & 96.82\% & 8.50\%  \\
&BERT  & 95.31\% & 94.37\% & \textbf{98.86}\% & 96.56\% & 11.80\%  \\
&Ours         & \textbf{95.98}\% &\textbf{100.00}\% & 94.00\% & \textbf{97.00}\% & 0.90\%  \\
\midrule
\multirow{5}{*}{\textbf{CHEAT}} 
&FastDetectGPT   & 54.33\% & 97.00\% & 9.00\%  & 16.00\% & \textbf{0.32}\% \\
&Binoculars      & 97.41\% & 95.49\% & \textbf{99.52}\% & 97.46\% & 4.70\%  \\
&GAN-BERT        & 83.56\% & 77.81\% & 93.90\% & 85.10\% & 26.80\% \\
&BERT  & 76.75\% & 68.46\% & 99.22\% & 81.02\% & 45.72\% \\
&Ours           & \textbf{97.83}\% & \textbf{98.00}\% & 98.00\% & \textbf{98.00}\% & 2.00\%  \\
\midrule
\multirow{5}{*}{\textbf{M4}} 
&FastDetectGPT   & 49.99\% & 0.00\%  & 0.00\%  & 0.00\%  & \textbf{0.02}\% \\
&Binoculars      & 71.35\% & \textbf{96.15}\% & 44.48\% & 60.82\% & 1.78\% \\
&GAN-BERT        & \textbf{76.12}\% & 77.44\% & \textbf{73.72}\% & \textbf{75.53}\% & 21.50\% \\
&BERT  & 66.42\% & 69.44\% & 58.66\% & 63.59\% & 25.82\% \\
&Ours          & 50.83\% & 66.00\% & 4.00\%  & 7.00\%  & 1.84\% \\
\bottomrule
\end{tabular}}
\caption{Results of all baselines on the in-domain RAID and out-of-domain (CHEAT, M4) benchmarks. Supervised methods outperform training-free approaches in-domain, while all models degrade under OOD shifts. Our method achieves the best performance on RAID and generalizes to CHEAT, but fails on M4 due to substantial domain mismatch, highlighting the difficulty of building a universal LLM detector across domains and models.}
\label{tab:unified_performance}
\end{table*}
Table~\ref{tab:unified_performance} summarizes results on the in-domain RAID dataset and the out-of-domain (OOD) CHEAT and M4 benchmarks. In-domain, supervised methods clearly outperform training-free approaches, with FastDetectGPT exhibiting extremely low recall, while Binoculars provides moderate gains. Among supervised models, GAN-BERT and BERT achieve strong performance, but our method attains the best overall results on RAID, achieving the highest accuracy and F1-score, perfect precision, and the lowest false positive rate. Under OOD evaluation, all methods show performance degradation. On CHEAT, our method generalizes well and achieves the best accuracy and F1-score with a low false positive rate, outperforming other supervised baselines. In contrast, on M4, performance drops sharply for all models; while GAN-BERT performs best among baselines, our method fails to generalize, with very low recall and F1. Overall, these results show that supervised detectors dominate in-domain settings, OOD shifts remain challenging, and achieving a universal LLM detector that generalizes across diverse domains and models remains difficult.

\paragraph{Why does our approach succeed on OOD CHEAT but fail on OOD M4?}
We analyze OOD generalization by evaluating models trained on RAID against the CHEAT and M4 benchmarks. Our framework transfers effectively to CHEAT, achieving 97.83\% accuracy, due to strong stylistic alignment between the datasets: the inclusion of ArXiv abstracts in RAID enables the style encoder to learn representations that generalize well to academic text. In contrast, performance drops sharply on M4 because of severe distributional shift. M4 exhibits substantially higher character diversity, increased digit density, and longer average text length compared to RAID, reflecting a move from formal sources such as Wikipedia and news to informal, noisy domains like Reddit \cite{wang2024m4}. This mismatch introduces linguistic variability that the learned style representations fail to capture. 
\begin{figure}[h]
    \centering
    \includegraphics[width=\columnwidth]{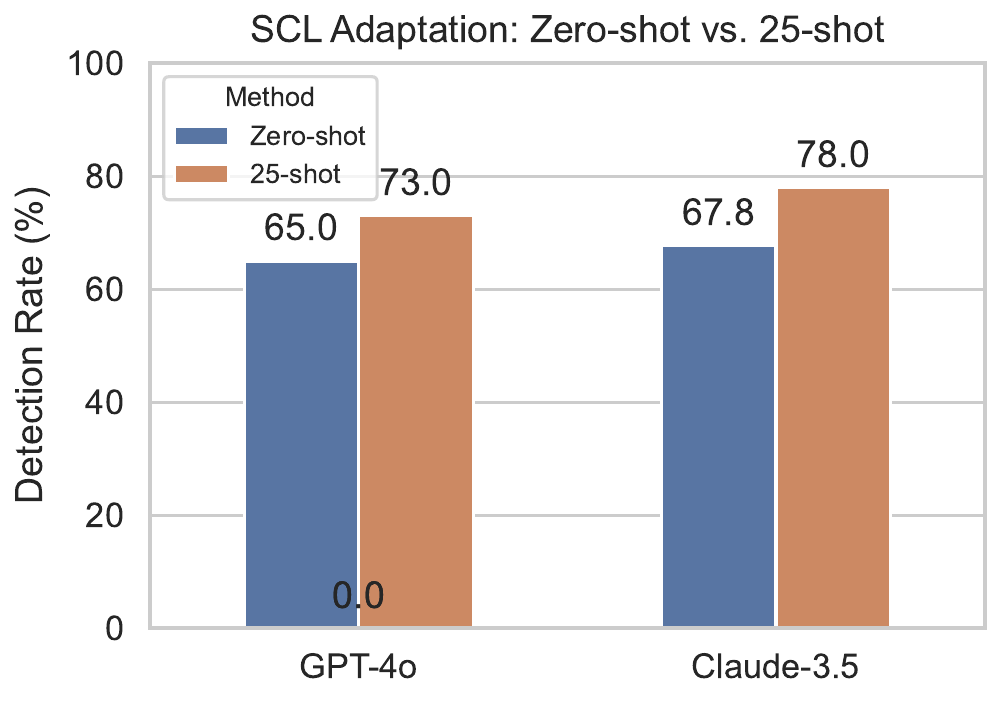}
    \caption{Zero-shot vs. 25-shot detection performance of our method on unseen LLMs, showing that lightweight adaptation without retraining yields substantial gains on both \texttt{GPT-4o} and \texttt{claude-3.5} LLMs.}
    \label{fig:scl_rates}
\end{figure}

\section{Analysis}
\subsection{Can our framework generalize to unseen LLMs and adapt with a few examples?}
We study whether our method can adapt to unseen LLMs (\texttt{GPT-4o} and \texttt{claude-3.5}) using only a few examples. Experiments are conducted on the LMSYS Arena dataset, which contains conversations from 55 chatbots and supports both zero-shot and few-shot generalization. We first compute class centroids by averaging style embeddings of human and AI samples. For adaptation to a new, unseen LLM, we generate a small number of model-specific samples (e.g., 25) and update only the AI centroid, without retraining the base model. Figure~\ref{fig:scl_rates} compares zero-shot and 25-shot performance, demonstrating that this lightweight adaptation yields substantial gains on both \texttt{GPT-4o} and \texttt{claude-3.5}. Overall, the results highlight the practicality and efficiency of the SCL framework for adapting to new LLMs.

\subsection{How robust is our proposed approach to adversarial perturbations?}
\label{sec:adversarial-robustness}
We evaluate the robustness of our framework under four adversarial settings. For white-box, gradient-based attacks, we assume full access to the detector and apply the Greedy Coordinate Gradient (\texttt{GCG}) attack \cite{zou2023universal}. We also test universal trigger attacks using a fixed adversarial suffix \cite{song-etal-2021-universal}, and consider simple black-box heuristic perturbations such as quotation marks, attribution cues, and typographical noise. 
White-box GCG attacks can flip predictions for each sample 99.3\%, but this success comes at the cost of semantic naturalness. It means the optimized texts exhibit broken syntax, repetition, and nonsensical tokens, making them easily identifiable by humans. In contrast, GCG universal attempts to find universal trigger for all samples but it fail to do so. Notably, the detector is most sensitive to simple black-box stylistic perturbations: adding quotation marks and attribution reduces accuracy from 3.3\% to 8.5\%, while introducing minor typos unexpectedly improves accuracy. These results indicate that current detectors rely on brittle surface cues, underscoring the need for more robust, style-aware detection methods.

\begin{table}[h!]
\centering
\begin{tabular}{@{}lc@{}}
\toprule
\textbf{Attack / Perturbation} & \textbf{Success} \\
\midrule
GCG (per-sample) &  99.3\%  \\
GCG (universal) &0\% \\
Quotation/Attribution & 8.5\%  \\
Typographical Noise &  3.3\% \\
\bottomrule
\end{tabular}
\caption{Results showing how different perturbations induce misclassification of AI-generated text as human. Per-sample GCG is highly effective, while universal GCG fails to generalize; simple noise such as quotation marks, attribution cues, and typographical perturbations also significantly impact performance.}
\label{tab:adv-summary}
\end{table}

\section{Related Work}
Although recent LLM detectors report near-perfect accuracy, they remain unreliable in practice, motivating analysis of their generalization limits of trustworthy detection. Existing approaches fall into supervised and training-free methods \cite{zhang2023survey}.
Early detectors exploited surface statistics such as n-grams, entropy, and perplexity, achieving strong in-domain performance (e.g., 99\% on CHEAT \cite{zhao2023cheat}) but exhibiting severe \emph{model brittleness} under generator, domain, and decoding shifts. Training-free methods aim to remove labeled data dependence by leveraging intrinsic text statistics. DetectGPT \cite{mitchell2023detectgpt}, along with Fast-DetectGPT \cite{bao2024fastdetectgpt} and Binoculars \cite{hans2024spotting}, improves efficiency and scalability, but detection remains highly sensitive to the proxy language model, limiting robustness under distribution shift.
Our experiments confirm that supervised detectors outperform training-free methods in-distribution but degrade substantially under OOD settings. Prior work addresses robustness via adversarial training (GAN-BERT \cite{croce2020ganbert}) and contrastive learning for OOD generalization \cite{aberdam2021sequence,chen2024spcontrastnet,guo2024detective}. Building on this, we introduce a contrastive objective into strong detectors and show that while robustness improves modestly, no universal detector generalizes reliably across models, domains, and adversarial strategies, revealing fundamental limits of current detection paradigms.

\section{Conclusion}
This work systematically examined whether current LLM detectors can be trusted under realistic deployment conditions. Through extensive in-domain, out-of-domain, and adversarial evaluations, we showed that both training-free and supervised detectors exhibit significant brittleness when confronted with unseen generators, domain shifts, or simple stylistic perturbations. While supervised methods dominate in-domain settings, their performance degrades sharply under distributional mismatch, and training-free approaches remain highly sensitive to proxy model choices. We introduced a supervised contrastive learning framework that improves robustness and enables efficient few-shot adaptation to new LLMs, demonstrating clear gains in certain OOD scenarios. Overall, our findings suggest that universal, domain-agnostic LLM detection remains infeasible with current paradigms.

\section*{Acknowledgments}
We thank Rishi Divyakirti, IIT Kanpur for helping with some of the experiments of this work. This work was supported by the “R\&D Hub Aimed
at Ensuring Transparency and Reliability of Generative AI Models” project of the Ministry of Education, Culture, Sports, Science and Technology

\bibliography{custom}

\end{document}